%% file: main_final.tex
\documentclass[10pt,twocolumn,letterpaper]{article}

\usepackage{iccv}
\usepackage{times}
\usepackage{epsfig}
\usepackage{graphicx}
\usepackage{amsmath}
\usepackage{amssymb}

\usepackage[accsupp]{axessibility}

\usepackage{times}
\usepackage{epsfig}
\usepackage{amsfonts}
\usepackage{nicefrac}       % compact symbols for 1/2, etc.https://www.overleaf.com/project/5ec714d0356a3d00015f1f63
\usepackage{algorithm}
\usepackage{algorithmic}
\usepackage{caption}
\usepackage{subcaption}
\usepackage{mathtools}
\usepackage{color}
\usepackage{setspace}
\usepackage{enumitem}
\usepackage{ctable}
\usepackage{soul}
\usepackage{multirow}
\usepackage{xspace}
\usepackage[toc,page]{appendix}
\usepackage[symbol]{footmisc}
\usepackage[breaklinks=true,bookmarks=false]{hyperref}

\iccvfinalcopy % *** Uncomment this line for the final submission

\def\iccvPaperID{6123} % *** Enter the ICCV Paper ID here

\ificcvfinal\fi

\newcommand{\twostream}[0]{two-stream }
\newcommand{\autotwostreamnetdense}[0]{Auto-TSNet}

\newcommand{\autotwostreamnetspace}[0]{Auto-TSNet }
\newcommand{\atn}[0]{Auto-TSNet }
\newcommand{\numvariables}[0]{6 }
\newcommand{\streamonelowercase}[0]{sparse }
\newcommand{\streamoneuppercase}[0]{Sparse }
\newcommand{\streamtwolowercase}[0]{dense }
\newcommand{\streamtwouppercase}[0]{Dense }
\newcommand{\kdata}[0]{Kinetics-400}

\begin{document}

%%%%%%%%% TITLE
\title{Searching for Two-Stream Models in Multivariate Space for Video Recognition}

% \and

\author{
  Xinyu Gong$^\dagger$$^\ddagger$\thanks{Work done during an internship at Facebook AI.} , Heng Wang$^\dagger$, Zheng Shou$^\dagger$, Matt Feiszli $^\dagger$, Zhangyang Wang$^\ddagger$, Zhicheng Yan$^\dagger$\thanks{Correspondence to Zhicheng Yan $<$zyan3@fb.com$>$.}\\
  $^\dagger$Facebook AI, $^\ddagger$The University of Texas at Austin \\
}

\maketitle
\ificcvfinal\thispagestyle{empty}\fi

%%%%%%%%% ABSTRACT
\input{tex/0_abstract.tex}

%%%%%%%%% BODY TEXT
% \footnote[1]{Work done during an internship at Facebook AI.}
\input{tex/1_intro}

\input{tex/2_related_work}
\vspace{-3pt}
\input{tex/3_search_space}

\input{tex/4_search_method}

\vspace{-3pt}
\input{tex/5_experiments}
\vspace{-6pt}
\input{tex/6_conclusion}

\clearpage

{\small
\bibliographystyle{ieee_fullname}
\bibliography{egbib}
}

% \clearpage 

% \appendix
% \renewcommand{\thefigure}{\thesection.\arabic{figure}}    
% \renewcommand{\thetable}{\thesection.\arabic{table}} 
% \begin{appendices}
% \input{tex/supp_1_search_detail}
% \input{tex/supp_2_fusion}
% \input{tex/supp_3_implementation}
% \input{tex/supp_4_latency}
% \end{appendices}

\end{document}

%% file: tex/0_abstract.tex
\begin{abstract}
% \zhicheng{ready to review}
 % \Heng{A very high level comment: Overall the paper seems to argue that two-stream network can learn spatial-temporal features because one stream focus on spatial and another stream focus on temporal. Also one stream network is not able to learn spatial-temporal features because it has only one stream. I am not sure whether the reviewer will agree to this, since they can argue that one stream network can also learn spatial-temporal features. Instead, if we want to justify that two-stream is better than one-stream, the argument could be: 1) two-stream network can learn complementary information, thus leads to better performance; 2) two-stream network has much large design space, thus leads to better FLOPs/accuracy trade-offs.}

% Spatial- and temporal features are both indispensable to action recognition in videos.\zs{remove the first sentence?} 

Conventional video models rely on a single stream to capture the complex spatial-temporal features. Recent work on two-stream video models, such as SlowFast network and AssembleNet, prescribe separate streams to learn complementary features, and achieve stronger performance. However, manually designing both streams as well as the in-between fusion blocks is a daunting task, requiring to explore a tremendously large design space. Such manual exploration is time-consuming and often ends up with sub-optimal architectures when computational resources are limited and the exploration is insufficient.  In this work, we present a pragmatic neural architecture search approach, which is able to search for two-stream video models in giant spaces efficiently.
We design a multivariate search space, including 6 search variables to capture a wide variety of choices in designing two-stream models.
Furthermore, we propose a progressive search procedure,
by searching for the architecture of individual streams, fusion blocks and attention blocks one after the other. We demonstrate two-stream models with significantly better performance can be automatically discovered in our design space.
Our searched two-stream models, namely Auto-TSNet, consistently outperform other models on standard benchmarks. On Kinetics, compared with the SlowFast model, our Auto-TSNet-L model reduces FLOPS by nearly $11\times$ while achieving the same accuracy $78.9\%$. On Something-Something-V2, Auto-TSNet-M improves the accuracy by at least $2\%$ over other methods which use less than 50 GFLOPS per video.

% by a considerable margin on Kinetics-400 and Something-Something-V2 benchmarks. Code and pre-trained models will be released upon acceptance.
%while using much less or an equal amount of FLOPS.   

\end{abstract} 

%% file: tex/1_intro.tex
\section{Introduction}

% \zhicheng{ready to review}
% \zs{instead of discussing spatio + temporal, directly start with an overarching statement about two-stream is needed/powerful, so that to better correspond to the motivation in 2nd paragraph -- our technique is to find better arch for two-stream (potentially could be any two streams, although in this paper we focus on the mainstream config of choosing these two streams)}

\begin{figure}[t!]
\vspace{-0.5cm}
   \centering
    %  \begin{subfigure}[b]{0.92\linewidth} 
    \includegraphics[width=1.0\linewidth]{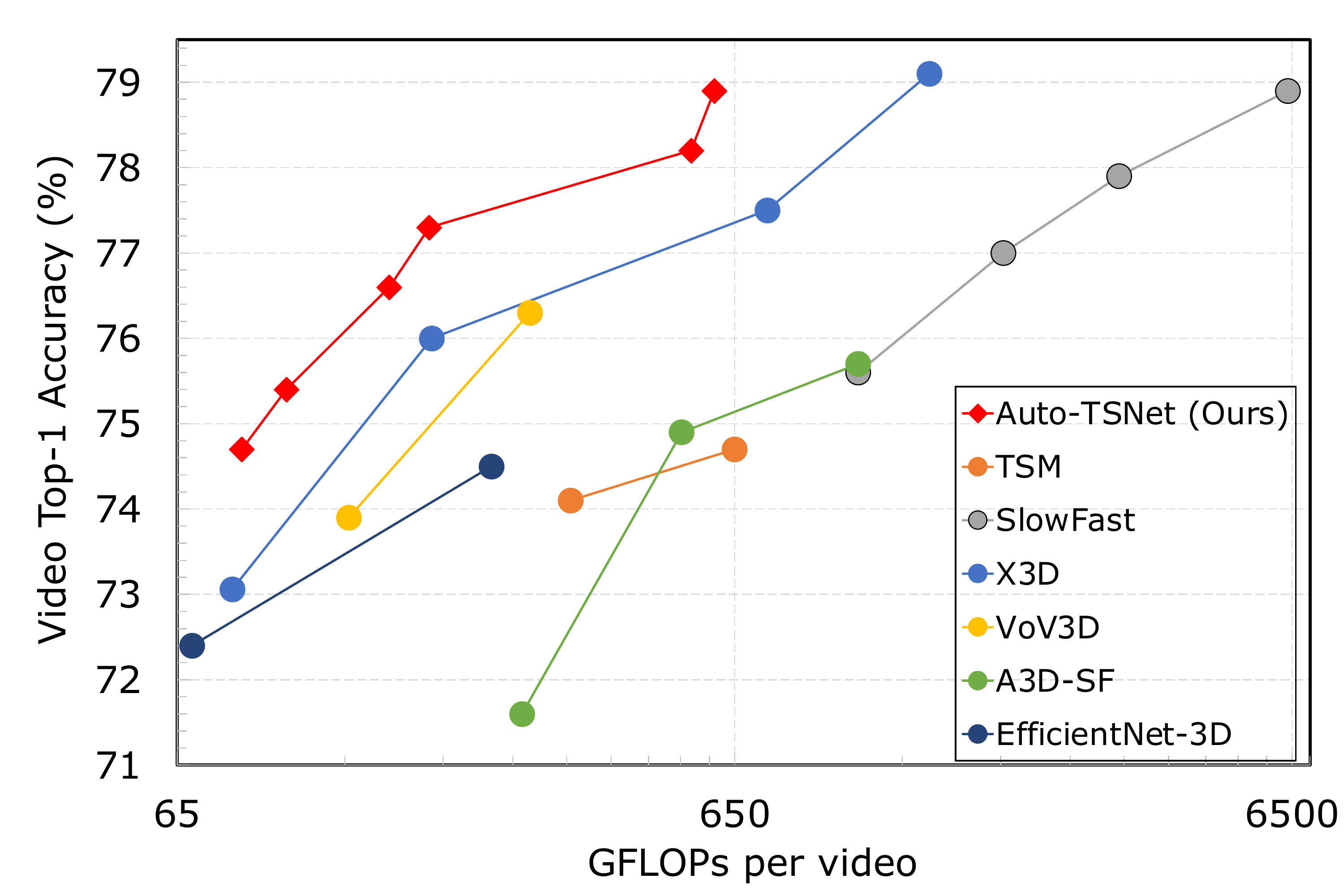}
    % \caption{Results on Kinetics-400 dataset.  \Xinyu{add more method data points.}}
    % \end{subfigure}
    \vspace{-10pt}
   \caption{\textbf{Results on Kinetics-400.} Comparing the FLOPs and accuracy with state-of-the-art models, our Auto-TSNet models achieve better accuracy-to-complexity trade-off. For a fair comparison, we report the FLOPs for each video at inference time, taking into account the different number of views used by each method.}
%   We report the total FLOPs for classifying the whole video since various methods sample a different number of views in each video. }
   \label{fig:sota_results}
\vspace{-20pt}
\end{figure}

Video recognition requires to learn both spatial and temporal features, which is arguably more challenging than image recognition. Many efforts have been made to extend single-stream image architectures for video recognition, such as C3D~\cite{c3d}, I3D~\cite{i3d}, S3D~\cite{s3d}, R(2+1)D~\cite{r2plus1}, TSN~\cite{temporalsegmentnetwork}, and TSM~\cite{tsm}. However, such single-stream models often underperform
two-stream models where each stream takes a separate input and learns spatial-temporal representations that are complementary to each other
~\cite{simonyan2014two, slowfast, twostream_residual}. 
In the pioneering two-stream ConvNet~\cite{simonyan2014two}, a separate temporal stream is added which takes multi-frame optical flow as input to better learn temporal information. Recently, SlowFast network~\cite{slowfast} adds a fast pathway, which operates at a high frame rate, and captures temporal information at a finer granularity. % temporal resolution.

Compared with the single-stream model, the number of design choices grows exponentially for the two-stream models as we need to take into account the additional complexity from the second stream, the feature fusion between the streams.
Prior hand-crafted two-stream models mitigate such challenges by largely reusing existing single-stream architectures, and only explore a limited number of customized design choices for each stream. For example, in two-stream ConvNet~\cite{simonyan2014two}, the second temporal stream shares the same architecture as the spatial stream, which doubles the overall computational cost of the model. In the SlowFast network~\cite{slowfast}, the fast pathway only differs from the slow pathway by using 1D temporal convolutions and uniformly reducing the feature channels to balance the accuracy-to-complexity (ATC) trade-off.
We hypothesize that the existing two-stream models are sub-optimal, and pose the following question: \textit{Can we more thoroughly explore the design space of two-stream video architectures and discover models with better performance? }

In this work, we present a pragmatic neural architecture search (NAS) approach, which can effectively explore large design spaces and discover high-performance \twostream models automatically. Unlike hand-crafted two-stream models where streams often use similar architectures, we encourage distinctive architectures at each stream, and jointly search both streams to learn complementary information. The core of our approach is a carefully prescribed multi-variate search space, which contains \numvariables search variables, including inter-stream fusion blocks, attention blocks, temporal/spatial kernel sizes, output channels, and expansion rates of building blocks. All of them have a substantial impact on both the accuracy and complexity of the learned model. Together they represent a wide variety of design choices for \twostream models.
% , and the resulting search space contains over $5 \times 10^{52}$ architectures.

It is computationally challenging to explore such gigantic search spaces efficiently.
We propose \textit{a multi-step progressive procedure} to decompose the large search space by only searching a smaller number of design choices at a time. 
For the basic search procedure, we adopt PARSEC~\cite{2019pnas} which is more memory efficient than other differential NAS methods by avoiding instantiating all choices of the search variables simultaneously, and only sampling one architecture at a time.  Unlike the process of manually designing architectures which often favors uniform choices of search variables, searching in our space leads to the discovery of the \textit{\autotwostreamnetdense}  models, which select \textit{more nonuniform choices} for different components of the models.
With extensive experiments, we demonstrate \autotwostreamnetspace models substantially outperform others with more uniform choices on Kinetics-400~\cite{kay2017kinetics}, shown in Figure \ref{fig:sota_results}, and Something-Something V2~\cite{sthsth} dataset. 

Our main contributions are summarized below.
\vspace{-5pt}
\begin{itemize}[leftmargin=*]
\item We prescribe a multi-variate search space to accommodate the large variations in designing \twostream video models, by including \numvariables different search variables, each of which has a significant impact on the model accuracy and complexity.
\vspace{-5pt}
\item We decompose the search of \twostream models into multiple steps, and sequentially search different parts of the model, which renders the exploration in such a large space more efficient. 
\vspace{-5pt}
\item The discovered \autotwostreamnetspace models are distinct from the hand-crafted ones by selecting more nonuniform choices for different components. 
We evaluate them on two large action recognition benchmarks, and confirm their superior performance over other models.
\end{itemize}
\vspace{-10pt}

%% file: tex/2_related_work.tex
\section{Related Work}
% \zhicheng{May need to condense related work}
\vspace{-6pt}
\noindent \textbf{One-Stream Video Models.} Video contains spatial-temporal signals and video recognition requires to extract both spatial and temporal features. One-stream video models, which are often built on top of image models, achieve such capability by various ways, such as replacing 2D with 3D convolution~\cite{c3d, i3d}, inserting 1D temporal convolution~\cite{r2plus1, s3d}, sampling temporal segments from the video~\cite{temporalsegmentnetwork}, and shifting feature channels along temporal dimension~\cite{tsm}. %Albeit substantial improvements over image models have been made for video recognition, single-stream models often underperform two-stream video models. \Heng{Maybe remove this? Or just add one introductory sentence for this in "Two-Stream Video Models.". }
% it is still challenging to use a single stream to capture the spatial semantics at different scales and temporal dynamics at different frequencies.
% \vspace{-4pt}

\noindent \textbf{From One-Stream to Two-Stream Video Models.} Since the defining difference between video and image is video contains the temporal information between frames, a number of two-stream models are proposed where an extra stream is dedicated to capture more temporal information complementary to that from the existing stream~\cite{3d_multiplier, twostreamconvnet, twostreampractice, wu2016multi}. The Two-Stream ConvNet~\cite{simonyan2014two} augments the single-stream model by feeding optical flow to a separate 2D stream. Two-Stream Residual Network~\cite{twostream_residual} improves it by introducing residual connections between streams. %Two-stream Fusion Network~\cite{twostream_fuse} improves the vanilla late fusion by fusing streams at earlier layers. TSM~\cite{temporalsegmentnetwork} uses two more streams to process RGB difference and optical flow. 
More recent SlowFast model~\cite{slowfast} employs slow and fast pathways to capture spatial semantics and temporal motion separately.  

%Beyond video recognition, two-stream models are also adopted for other tasks, such as video super-resolution~\cite{twostream_vsr}, video object segmentation~\cite{Fusionseg}, action detection~\cite{twostream_action_detect}, and compressed video recognition~\cite{wu2018compressed}. However, aforementioned two-stream models often borrow the architecture from existing 2D/3D models, using similar architectures in both streams, and only manually explore a limited set of choices.
% \vspace{-4pt}
\noindent \textbf{Neural Architecture Search.} NAS methods automatically search models in a predefined space, and the searched 2D models has already surpassed the hand-crafted ones. NAS methods can be based on RL~\cite{2019Mnasnet, 2019efficientnet, 2019nasnet}, evolution~\cite{2017large-scale-evolution, 2019AmoebaNet}, and differentiable search~\cite{2018darts, 2019atomnas, 2019pnas}. NAS has also been used to search video models. CAKES~\cite{yu2020cakes} searches channel-wise spatial/temporal kernels to improve model efficiency. 
%while EvaNet~\cite{piergiovanni2019evolving} searches over 6 types of space-time convolution/pooling layers. 
X3D~\cite{x3d} gradually and uniformly expands a 2D model along 6 dimensions (\eg, spatial resolution, model width) to derive efficient 3D models. 

The design space of two-stream video models is significantly larger, and prior efforts have focused on searching fewer variables.  AssembleNet~\cite{assemblenet} only searches the connections between streams, while keeping the architecture of building blocks fixed.  AssembleNet++~\cite{assemblenet++} introduces the new object stream, but only searches the block connectivity for SqueezeExcite~\cite{2018senet} module. Different from prior work, we search two-stream video models over \numvariables different variables, which is crucial for improving the ATC trade-off of the model. Our design space allows different architectures in individual streams and nonuniform design choices over different parts of the stream, and captures the variations in inter-stream fusion blocks, attention blocks, and the stream architectures. A progressive search process is proposed to efficiently search in such large spaces.

%% file: tex/3_search_space.tex
\section{Two-Stream Multivariate Search Space}
\label{sec:search_space}

% \zs{organzie this section into subsections: overview/motivation, MBConv3D, Fusion, Attention}
\vspace{-6pt}
\input{tex/3_1_space_overview}
\vspace{-6pt}
\input{tex/3_2_two_stream_fusion}
\vspace{-6pt}
\input{tex/3_3_two_stream_attention}

\vspace{-6pt}
\input{tex/3_4_searchable_mbconv_block}

\vspace{-6pt}
\input{tex/3_5_the_final_space}

\vspace{-6pt}

%% file: tex/3_1_space_overview.tex
\subsection{Overview}\label{overview}

\begin{figure}[t!]
    
    \begin{subfigure}[b]{\linewidth} 
    \centering
    \includegraphics[width=0.85\linewidth]{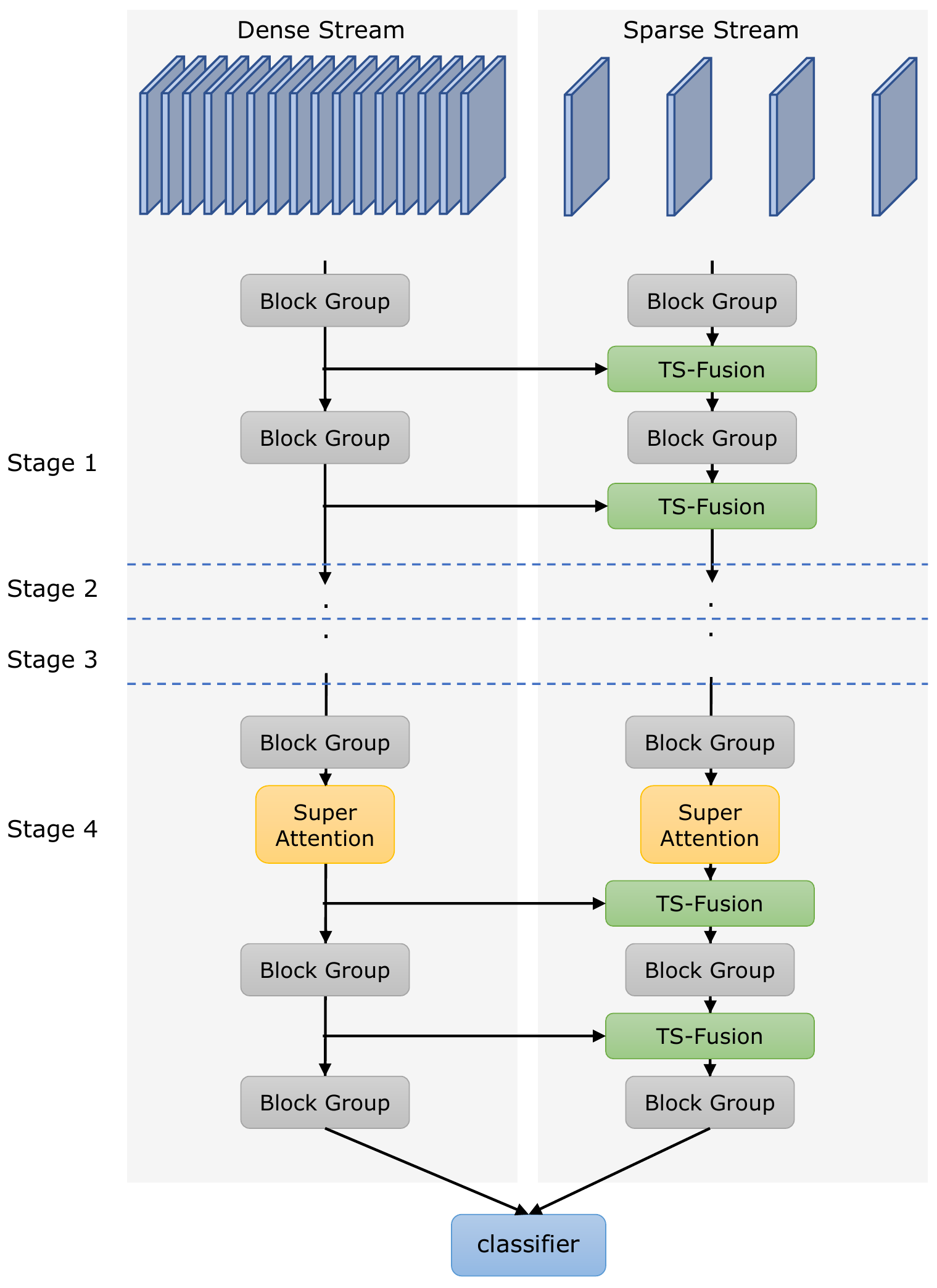}
    \caption{Multi-Variate Two-stream Super Model}
    \label{fig:search_space_a}
    \end{subfigure}
    
    \begin{subfigure}[b]{\linewidth} 
    \centering
    \includegraphics[width=0.9\linewidth]{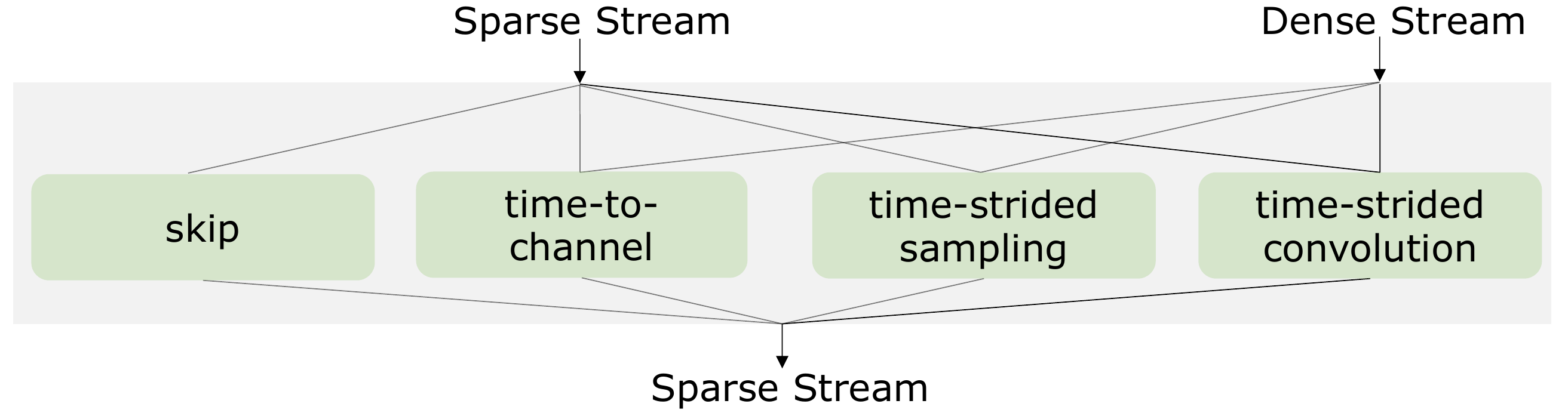}
    \caption{Super TS-Fusion Block}
    \label{fig:search_space_b}
    \end{subfigure}
    
    % \begin{subfigure}[b]{\linewidth} 
    % \includegraphics[width=1\linewidth]{figs/ts_attention}
    % \caption{Super TS-Attention Block}
    % \end{subfigure}
    
    \begin{subfigure}[b]{0.9\linewidth}
     \centering
    \includegraphics[width=\linewidth]{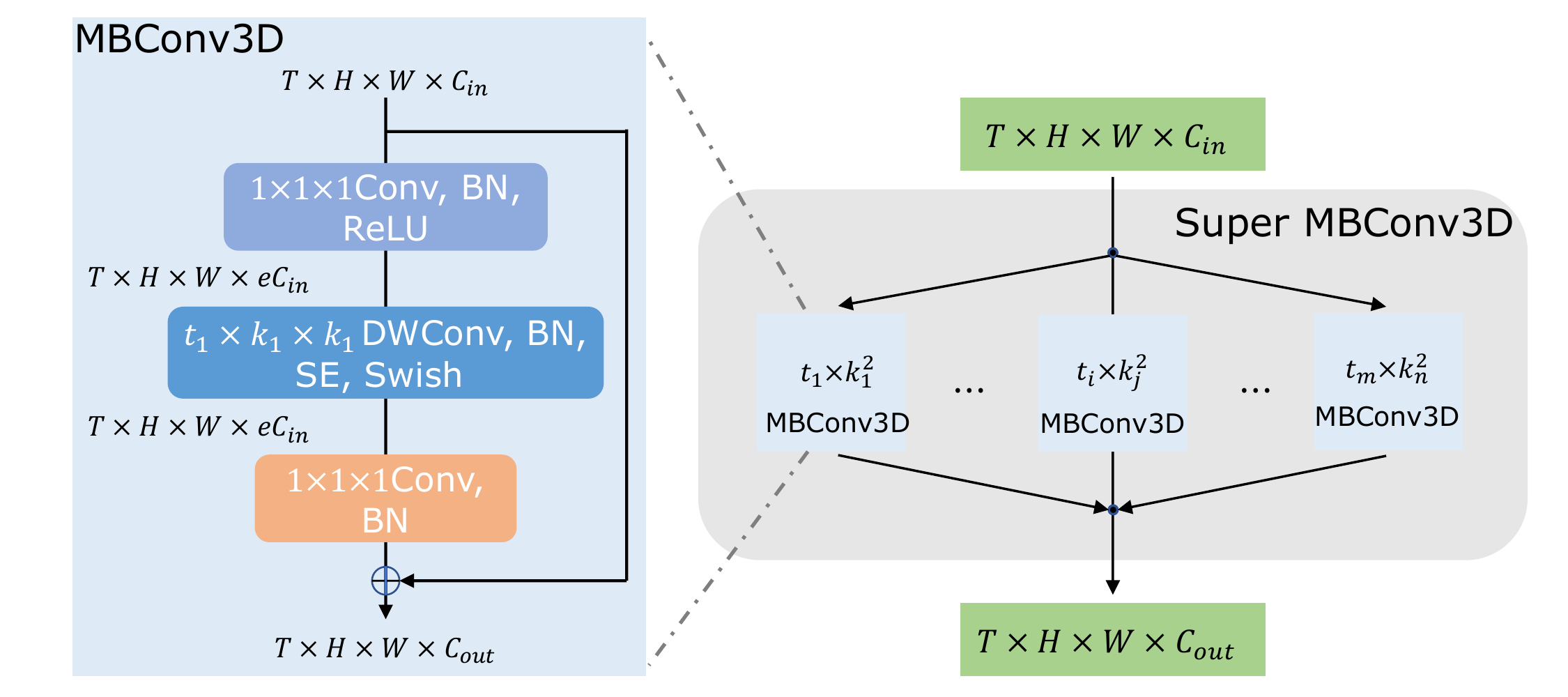}
    \caption{Super MBConv3D Block}
    \label{fig:search_space_c}
    \end{subfigure}
    \vspace{-5pt}
  \caption{
  \textbf{The multi-variate search space of two-stream models.} \textbf{(a)} The macro-architecture, where we define the layout of the various block groups in the model. \textbf{(b)} Super TS-Fusion block, where we search for the type of fusion operation. \textbf{(c)} Super MBConv3D block, where we search for the kernel size ($t$ and $k$) of 3D depthwise convolution, the output channel $C_{out}$, and the expansion rate $e$.}
  \vspace{-8pt}
  \label{fig:search_space} 
\end{figure} 
% Inspired by the pioneering hand-crafted two-stream models~\cite{twostreamconvnet, slowfast}, w
We define a multi-variate search space of two-stream models which use separate streams to capture complementary spatial-temporal signals. As shown in Figure~\ref{fig:search_space_a}, a \textit{\streamonelowercase stream} takes sparse frames as input which are sampled from the video with a larger temporal stride, while a \textit{\streamtwolowercase stream} takes dense frames with a smaller temporal stride. The proposed search space can be decomposed into three parts. 

\noindent \textbf{Fusion}. To learn complementary signals between two streams, we allow to fuse features from two streams at different layers of the model. Rather than manual designing, we search for the fusion operations and their locations between the two streams.
% \vspace{-3pt}

\noindent \textbf{Attention}. Attention block has been used to improve the accuracy~\cite{nonlocal_block, dual_attention, woo2018cbam}. % network performance
Previous work has shown that the design of the attention blocks and their placement locations are critical to the final performance. 
We propose to search for these design choices to yield more competitive models.
%the number \& location of attention blocks have a large impact on the network performance. 
%Therefore, we also search for the number and location of attention blocks in the backbone.
% \vspace{-3pt}

\noindent \textbf{Backbone}. The model backbone comprises stacked building blocks and represents a significant portion of the computational cost. %the model computation. 
% As one of the most prominent components of a neural network, most NAS papers focus on searching for a powerful backbone. 
In our work, we adopt a hierarchical backbone search space. As shown in Figure \ref{fig:search_space_a}, each stream of the network backbone includes 4 stages, and each stage has multiple block groups, where blocks in the same group share the same architecture. 

%In total, we define \numvariables search variables in the space, and present more details for each one below.

%% file: tex/3_2_two_stream_fusion.tex
\subsection{Searchable Two-Stream Fusion Block}
\vspace{-5pt}
Early two-stream models~\cite{twostreamconvnet, temporalsegmentnetwork} do not fuse features from individual streams in the middle of the model. Recent SlowFast work~\cite{slowfast} adds a fusion block to fuse features from both streams at the end of each stage, and uses the same type of fusion operation at all places. %model chooses the end of each stage where a fusion block is added to fuse features from streams, and also a uniform type of fusion operation. 
% Even with such simple manual design, SlowFast has shown significant gain in the recognition accuracy. 
We hypothesize such uniform design choice is sub-optimal, and %more sufficiently 
propose to explore the design choices of how many fusion blocks to add, where to add and what type of fusion to use.
% and automatically search the fusion blocks as follows.
At the end of each group, we add a \textit{super two-stream fusion block}, as shown in Figure~\ref{fig:search_space_b}, where we search for the type of fusion operation for combining the features from two streams and passing the fused feature to the next block in the \streamonelowercase stream. The candidate fusion operations are adopted from~\cite{slowfast}, and details are included in the supplement.

\begin{table*}[t!]
% \vspace{-0.6cm}
\centering
% \small
% \tiny
% \scriptsize

\subfloat[\streamoneuppercase stream]{

\resizebox{0.51\textwidth}{!}{

  \begin{tabular}{c|cccccc|cc}

    Max Input size & \multicolumn{6}{c}{Block group} &Fusion& Attention \\
     $C \times T \times S^2$ &Stage& Operator & Channels   & Expansion & N &  S &Block& Block \\
    \midrule
    % \specialrule{.15em}{.1em}{.1em}

% stem
    $3 \times 4 \times 224^2$ &\multirow{2}{*}{0}& conv $1\times3^2$ & 24 &\multirow{2}{*}{-}& 1  &   2 &\multirow{2}{*}{-}&  \multirow{2}{*}{-}   \\ %\hline
    $3 \times 4 \times 112^2$  &&  conv $3\times1^2$ & 24 & & 1  &   1 && \\\hline
    
% stage 0
  $48 \times 4 \times 112^2$   &\multirow{2}{*}{1}& \multirow{2}{*}{MBConv3D} & (32, 48, 8)    & \multirow{2}{*}{(1.5, 6.0, 0.75)}      & 1 & 2 &\multirow{2}{*}{TS-Fusion}&  \multirow{2}{*}{-}  \\ 
  
  $48 \times 4 \times 56^2$   &&  & (32, 48, 8)  &  & 2 & 1 &&\\ \hline

% stage 1
  $88 \times 4 \times 56^2$  &\multirow{2}{*}{2}& \multirow{2}{*}{MBConv3D} & (64, 88, 8) &  \multirow{2}{*}{(1.5, 6.0, 0.75)}  & 1  & 2 &\multirow{2}{*}{TS-Fusion}&  \multirow{2}{*}{GloRe}\\ 
  
  $88 \times 4 \times 28^2$  && & (64, 88, 8) &  & 4 & 1 && \\ \hline

% stage 2
  $176 \times 4 \times 28^2$  &\multirow{4}{*}{3}& \multirow{4}{*}{MBConv3D} & (128, 176, 16) & \multirow{4}{*}{(1.5, 6.0, 0.75)}       & 1 & 2 &\multirow{4}{*}{TS-Fusion}&  \multirow{4}{*}{GloRe}  \\ 
  
  $176 \times 4 \times 14^2$   && & (128, 176, 16) &         & 3 & 1 && \\

  $176 \times 4 \times 14^2$   && & (128, 176, 16) &          & 3 & 1 && \\

  $176 \times 4 \times 14^2$   && & (128, 176, 16) &      & 4 &  1 && \\ \hline

% stage 3
  $344 \times 4 \times 14^2$ &\multirow{3}{*}{4}& \multirow{3}{*}{MBConv3D} & (248, 344, 24) & \multirow{3}{*}{(1.5, 6.0, 0.75)}   & 1 & 2  &\multirow{3}{*}{TS-Fusion}&  \multirow{3}{*}{GloRe}\\

  $344 \times 4 \times 7^2$  &&  & (248, 344, 24) &          & 3 & 1 && \\
  
  $344 \times 4 \times 7^2$   &&  & (248, 344, 24) &          & 3 & 1 && \\
  
%   312\times7^2 & conv 1\times1\times1 & - & 432 & 1 & 1 \\
  
%   432\times7^2 & avgpool & - & - & 1 & 1 \\
%   432 & fc & - & 2048 & 1 & -  \\
%   2048 & fc & - & \#classes & 1 & -  \\
\bottomrule
% \specialrule{.1em}{.05em}{.05em}

  \end{tabular}
  
}

}\quad
\subfloat[\streamtwouppercase stream]{

\resizebox{0.44\textwidth}{!}{

  \begin{tabular}{c|cccccc|c}
    Max Input size & \multicolumn{6}{c}{Block group} & Attention\\
     $C \times T \times S^2$ &Stage& Operator & Channels  & Expansion & N & S & Block \\ \midrule
    % \specialrule{.15em}{.1em}{.1em}

% stem
    $3 \times 32 \times 224^2$ &\multirow{2}{*}{0}& conv $1\times3^2$ & 8 & \multirow{2}{*}{-} & 1  &  2 &  \multirow{2}{*}{-} \\ 
    $3 \times 32 \times 112^2$ &&  conv $3\times1^2$ & 8 &  & 1  &  1 &  \\\hline
% stage 0
  $8 \times 32 \times 112^2$ &\multirow{2}{*}{1}& \multirow{2}{*}{MBConv3D} & 8    & \multirow{2}{*}{(1.5, 6.0, 0.75)}      & 1 & 2&  \multirow{2}{*}{-}    \\ 
 $16 \times 32 \times 56^2$ &&  & (8, 16, 8)      &    & 2 & 1 & \\ \hline
  
% stage 1
  $24 \times 32 \times 56^2$ &\multirow{2}{*}{2}& \multirow{2}{*}{MBConv3D} & (8, 24, 8)      &  \multirow{2}{*}{(1.5, 6.0, 0.75)}  & 1 & 2 &  \multirow{2}{*}{GloRe}\\ 
  
  $24 \times 32 \times 28^2$ &&  & (8, 24, 8)    &     & 4 & 1 & \\  \hline

% stage 2
  $32 \times 32 \times 28^2$  &\multirow{4}{*}{3}&\multirow{4}{*}{MBConv3D} & (16, 32, 8) &  \multirow{4}{*}{(1.5, 6.0, 0.75)}  & 1 & 2 & \multirow{4}{*}{GloRe}\\

  $32 \times 32 \times 14^2$  && & (16, 32, 8) &       & 3 & 1 &  \\

  $32 \times 32 \times 14^2$  && & (16, 32, 8) &        & 3 & 1 & \\

  $32 \times 32 \times 14^2$  && & (16, 32, 8) &    & 4 & 1 & \\  \hline

% stage 3
  $56 \times 32 \times 14^2$ &\multirow{3}{*}{4}&\multirow{3}{*}{MBConv3D} & (32, 56, 8) & \multirow{3}{*}{(1.5, 6.0, 0.75)} & 1 & 2 &\multirow{3}{*}{GloRe} \\

  $56 \times 32 \times 7^2$  && & (32, 56, 8) &          & 3 & 1 & \\
  
  $56\times 32 \times 7^2$   && & (32, 56, 8) &          & 3 & 1 & \\ 
  
\bottomrule
% \specialrule{.1em}{.05em}{.05em}

  \end{tabular}
  
}

}

\vspace{-5pt}
\caption{\textbf{Two-stream macro architecture}. \textbf{(a)} The macro-architecture of the \streamonelowercase stream. Each row represents a block group, which includes multiple MBConv3D blocks with the same architecture. It also has a searchable GloRe attention block and a TS-Fusion block at the end. Columns \textit{Channel} and \textit{Expansion} denote the output channel and the expansion rate of the block. Their search choices are denoted as a range \textit{(min, max, step)}. Column \textit{N} denotes the repeating times of the MBConv3D block, and column \textit{S} is the spatial stride of first block in the group. \textbf{(b)} The macro-architecture of the \streamtwolowercase stream.}
\label{tab:macro-structure}
\vspace{-6pt}
\end{table*}

%% file: tex/3_3_two_stream_attention.tex
\subsection{Searchable Attention Block}
\vspace{-5pt}
 Attention blocks, such as Non-Local~\cite{nonlocal_block} and GloRe blocks~\cite{glore}, can be inserted into the backbone to improve the performance. 
%  One has to carefully choose the location to insert the attention block and number of attention blocks.
In the case of manual design, one has to carefully decide where to add attention blocks and how many to add~\cite{nonlocal_block,glore}. Different choices have a large impact on both the model recognition performance and model FLOPS.
Therefore, we search for the number and location of attention blocks, in order to achieve a better ATC trade-off.

 In particular, we choose to use GloRe~\cite{glore} as an instantiation of the attention block.  Other attention blocks, such as non-local block~\cite{nonlocal_block}, can also be used.
In our search space, \textit{searchable attention blocks} are placed between block groups, which is denoted as Super Attention in Figure~\ref{fig:search_space_a}. We decide whether the features from each stream should simply pass through it or enter the attention block to perform global reasoning. % separately. 

%% file: tex/3_4_searchable_mbconv_block.tex
\subsection{Searchable Backbone Building Block } 

% \zhicheng{need to condense this section. The experiments section should start no later than 4.5 page}
We design a hierarchical search space, where the backbone is composed of block groups of different architecture (Figure \ref{fig:search_space_a}). Each group consists of several blocks, which are stacked sequentially and share the same architecture. 
In this work, we adopt a 3D version of MBConv (Mobile inverted Bottleneck Conv), originating from MobileNetV2~\cite{2018mobilenetv2}, namely \textbf{MBConv3D} (Figure~\ref{fig:search_space_c}), which has 4 search variables, including temporal \& spatial kernel size $t$ \& $k$, output channel $C_{out}$ and expansion rate $e$. %which are illustrated below. 
%We introduce 4 more variables to search for MBConv3D block below.

\noindent{\textbf{Temporal kernel of depthwise convolution.}} Temporal kernel size has a large impact on the model FLOPS and accuracy. When it is set to 1, the MBConv3D block only has 2D convolution, which is computationally cheaper but is not able to capture temporal information. When it is larger than 1, the MBConv3D block is doing convolution in 3D, which is more costly but can improve the recognition performance by capturing temporal signals. %In previous hand-crafted models~\cite{s3d, slowfast}, the temporal kernel size at different layers is delicately manually tuned to strike a balance between accuracy and complexity. %For example, in S3D model~\cite{s3d} and the slow pathway of SlowFast model~\cite{slowfast} where the model has 4 stages, they choose to use 2D convolutions in the first 2 stages, and 3D convolution with uniform temporal kernel size 3 in the last 2 stages to strike a balance between accuracy and complexity. 
We search the temporal kernel size for each block group. This not only avoids the tedious manual tuning but also improves the ATC trade-off.

\noindent{\textbf{Spatial kernel of depthwise convolution.}} Spatial kernel size is always considered to be critical for the model complexity and performance~\cite{2019Mnasnet, fbnet}. In hand-crafted video models, such as I3D~\cite{i3d}, S3D, SlowFast, and X3D, the spatial kernel is fixed to be 3 in almost all convolutional layers. We hypothesize such simple choice is sub-optimal, and add spatial kernel size to our search space as well. %search for the spatial size for each individual group to improve the ATC trade-off.

\noindent{\textbf{Output channel of MBConv3D block.}} In X3D~\cite{x3d}, which also adopts MBConv3D block as the building block, the output channel of each block is manually prescribed following simple heuristics, such as the output channel number doubles when the spatial resolution is reduced by half. As the choices of output channels substantially impact both the computation cost and capacity of the model, % block's capability of transforming features, 
we search for the output channel from a wider range of choices.

\noindent{\textbf{Expansion rate of the MBConv3D block.}} 
% \Heng{maybe add one sentence to explain the definition of expansion rate for readers that are not similar with it.} 
A MBConv3D block expands the feature channel through a point-wise convolution by an expansion rate, performs 3D depth-wise convolution, and finally shrinks the feature channel through another point-wise convolution. The expansion rate affects the number of feature channels where the depth-wise convolution operates, and consequently the ATC trade-off of the model.
Previous models~\cite{2018mobilenetv2, x3d} uses a constant expansion rate for all blocks. In contrast, we search for a separate expansion rate for each block.
%Such a uniform choice is not expected to be optimal. %We show in the experiments that by searching the expansion rates for individual groups of MBConv3D blocks, better ATC trade-off can be achieved.

%% file: tex/3_5_the_final_space.tex
\subsection{The Final Search Space}
\vspace{-3pt}

\begin{table}[t!]
% \vspace{-0.6cm}
\centering
% \small
% \tiny
% \scriptsize
\resizebox{0.4\textwidth}{!}{
  \begin{tabular}{c|cc}
    % &   & \multicolumn{2}{c}{Search Variables}  \\
    Search variable & Temporal kernel & Spatial Kernel  \\
%     \specialrule{.15em}{.1em}{.1em}
\midrule
    Choices & $\{1, 3, 5\}$ & $\{3, 5\}$ \\
  \end{tabular}
}
\vspace{-3pt}
\caption{\textbf{The choices of kernel size for MBConv3D block.}}
\label{tab:micro-structure}
\vspace{-10pt}
\end{table}

The full specification of the macro architecture of our search space is shown in Table~\ref{tab:macro-structure}. Our search space has \numvariables variables (temporal \& spatial kernel size, channel width, expansion rate, fusion, and attention block), where each block group can have its unique choice. It contains over $2\times10^{53}$ architectures, and poses a great challenge to efficiently search architectures within it. The choices of temporal and spatial kernel sizes for the MBConv3D block are shown in Table~\ref{tab:micro-structure}.

%% file: tex/4_search_method.tex
\section{Search Method}
\label{sec:progressive_search}

% \zhicheng{ready to review}

%\Heng{We design efficient search algorithms to fully explore our design search space}

% \zhicheng{Wrong arrangement. We should present PARSEC basics first, progressive search second, and cost-aware search in the last.} \Xinyu{I think we can merge cost-aware with the basic PARSEC. Since it is only an additional regularization term in addition to the search target.}

\input{tex/4_0_PARSEC}
\vspace{-4pt}
\input{tex/4_1_progressive_search}

% \input{tex/4_2_cost_aware}

%% file: tex/4_0_PARSEC.tex
\subsection{Background of Search Algorithm}

We adopt the PARSEC~\cite{2019pnas} method as our basic search procedure, which is a probabilistic version of the differentiable NAS method DARTS~\cite{2018darts}. Unlike DARTS, which requires to simultaneously instantiate all the layer choices and has a large memory footprint, PARSEC only samples one architecture at a time, and uses the same memory as in the standard model training.  

PARSEC constructs a supernet where we can sample architectures $\{A_i\}$ according to a distribution $ P(A|\pmb{\alpha})$. Architecture parameters $\pmb{\alpha}$ denote the probabilities of choosing different operations. Leveraging Importance Weighted Monte-Carlo algorithm~\cite{carlin2000empirical}, we jointly optimize $\pmb{\alpha}$ and model weights of the supernet to maximize the data likelihood of sampled architectures, which are weighted by a proxy architecture performance indicator (video-level accuracy on validation set in our paper).  We also add a hinge-type regularization term in the loss function to penalize architectures exceeding the target FLOPs range. 

%% file: tex/4_1_progressive_search.tex
\subsection{Progressive Search}

Directly searching for the architecture of all parts of the two-stream model in our large search space is challenging. Therefore, we consider a divide-and-conquer strategy by breaking down the search process into multiple steps, and searching for different parts of the model sequentially. Empirically, we found such progressive procedure can accelerate searching efficiency without sacrifice on the performance of the final discovered architecture, compared to searching for all model parts simultaneously. 
% Further details and discussions are included in the supplement.

% \Heng{Do we want to write this as an Algorithm? We can further break down each step.}

\noindent{\textbf{Step 1:}} We first search the architecture of the \streamonelowercase stream, including temporal/spatial kernel, the output channel, and expansion rate of the MBConv3D blocks, which lives in a greatly reduced search space of size $8\times10^{24}$.

\noindent{\textbf{Step 2:}} After that, we fix the architecture of the \streamonelowercase stream and inherit the model weight from step 1, and further search the architecture of the \streamtwolowercase stream as well as the fusion blocks to optimize the performance of \textit{the overall \twostream model}. Therefore, the search of \streamtwolowercase stream and fusion blocks favors the architectures which are more capable of learning complementary features from two streams. The search space in this step contains $6\times10^{24}$ unique architectures.

\noindent{\textbf{Step 3:}} In the final step, we search for the location to add the attention blocks which can improve the performance of the overall \twostream model at low computational cost, while keeping previous searched architecture fixed and inheriting their model weights. The search space in this step has a small size of 4096.

% Our final searched model consists of a \streamonelowercase stream, a \streamtwolowercase stream, multiple fusion blocks and attention blocks. The architectures of all of them are obtained by search.  In Section~\ref{sec:exp_search_two_stream_model}, we show such progressive search process can discover high-performance two-stream models at affordable search cost (256 GPU days), and ensure the two streams learn complementary spatial-temporal representations as the search in the following steps is conditioned on the discovered architecture of the model parts from previous steps.

% In DARTS search space, which only contains $3\times10^{18}$ architectures and is much smaller than ours, the original PARSEC samples 16 architectures at each optimization step, and run forward/backward pass on all of them. As the search space grows as large as ours, the architecture sample size needs to increase which will significantly increase the search time. As we empirically show in Section~\ref{sec:exp_search_two_stream_model}, naively searching in our giant search space is slow. 

%% file: tex/5_experiments.tex
% \vspace{-6pt}
\section{Experiments}

\label{sec:experiments}
\vspace{-3pt}
\input{tex/5_0_datasets}
\input{tex/5_1_setup}
\vspace{-3pt}
\input{tex/5_2_main_results}
\input{tex/5_3_twostream_baseline}

\input{tex/5_5_search_two_stream_model}
\vspace{-3pt}
\input{tex/5_6_component_ablation}
\vspace{-3pt}
\input{tex/5_6_progressive_ablation}
\vspace{-3pt}
\input{tex/5_7_sota_models_kinetics}

\vspace{-4pt}

\input{tex/5_8_sota_models_sth_sth}
\vspace{-3pt}
\input{tex/5_9_Analysis_of_model}

%% file: tex/5_0_datasets.tex
\subsection{Datasets}

We use two large-scale video 
benchmarks Kinetics-400~\cite{kay2017kinetics} and Something-Something-V2~\cite{sthsth}, which capture different aspects of video recognition tasks.
%are used to evaluate our \autotwostreamnet models. 
\textit{(i) Kinetics-400:} It contains 240K training- and 20K validation trimmed videos in 400 action classes, and focuses on general-purpose action recognition.
%We report the top-1/top-5 accuracy on its validation set. We also use Kinetics for ablative studies. 
\textit{(ii) Something-Something-V2:} Unlike Kinetics, it decouples human actions and the objects involved in the actions, and forces the model to learn temporal information instead of recognizing objects.
%Compared with Kinetics, it requires stronger model capability to capture the temporal signal, as shown in prior work~\cite{s3d, zhou2018temporal}. 
It contains 169K and 25K videos in the training and validation set from 174 human action classes. 
Our proposed \autotwostreamnetspace can generalize to datasets of different characteristics and achieve superior performance.
Following the standard protocol, we report the top-1/top-5 validation accuracy for both datasets.
%We report our result on  Something-Something-V2 in our supplementary material \zhicheng{we should report results on sth-sth-v2 in the 8-page paper} . 

%% file: tex/5_1_setup.tex
%\subsection{Settings}
\subsection{Implementation details}
% \Xinyu{Will move some to supp if space is limited.}
% \noindent{\textbf{Architecture Search}} We randomly select 100 classes in the Kinetics, denoted as \textit{MiniKinetics-100}, and search architectures on this smaller dataset for fast search.  We search for the architecture of \streamonelowercase stream for 800 epochs, where we use the first 400 epochs to warm-up the supernet, i.e., freeze the architecture parameters and only update the model weights. The search for \streamtwolowercase stream takes 400 epochs, including 100 warm-up epochs. The locations of attention blocks are searched for 200 epochs, containing 50 warm-up epochs. Adam optimizer is adopted to update the architecture parameters, with 0.025 learning rate and 0 weight decay. Model weights in the supernet are optimized with SGD, whose learning rate are set to 0.4 with a schedule of cosine decaying.
We briefly introduce the searching, training and evaluation setup below, and include more details in the supplement.

\noindent{\textbf{Architecture Search}} We randomly select 100 classes from Kinetics-400 dataset, denoted as \textit{MiniKinetics-100}, and search architectures on it for fast search. The architecture of \streamonelowercase stream is searched for 800 epochs. The search for \streamtwolowercase stream and the locations of attention blocks takes 400, 200 epochs respectively. Adam optimizer is adopted to update the architecture parameters, with learning rate 0.025  and zero weight decay. Model weights in the supernet are optimized with SGD, which uses learning rate 0.4 with a schedule of cosine decaying.

The \streamonelowercase and \streamtwolowercase stream take 4 and 32 frames as input, respectively. We use a scale jittering range of [182, 228] and then take a random crop of size $160 \times 160$ from each frame of the input video. %The frame rate of the input video  for \streamonelowercase stream is set to 12, and the \streamtwolowercase stream takes the video of 1.5 frame rate as input accordingly.
%By default, we use 64 Nvidia V100 GPU in our experiments.

\noindent{\textbf{Training the searched models.}} After the search, we take the most probable architecture and \textit{train it from scratch with model weights randomly initialized}. %We adopt the open source ClassyVision~\cite{classyvision}, which is a PyTorch framework for video classification. 
We train the model for 300 epochs. We use SGD optimizer, and learning rate 0.4 with a schedule of cosine decaying.
%, where the first 34 epochs are used for warming up learning rate. We set momentum and weight decay to 0.9 and $0.0005$, respectively. 
%The dropout rate is set to 0.5 at the network head. 

% The data processing part in training phase is identical to the search phase. 

%On Something-Something-V2, we use the pretrained model weights on Kinetics-400 for initialization. The model is finetuned for 60 epochs, with learning rate 0.6. Other settings are identical to those in the training on Kinetics-400.

\noindent{\textbf{Evaluating the searched models.}} The trained model is evaluated on the validation set. We uniformly sample $10$ clips from each video, and use two different ways of taking crops to obtain results comparable to those from prior work. \textit{(1) 10-Center:} we resize the clip to have a short edge 182. A single central crop of size $160^2$ is taken. \textit{(2) 10-LeftCenterRight:} we take 3 crops of size $182^2$ to cover the longer axis of the clip. The predictions is averaged over all crops of the clips. By default, our results are obtained by 10-Center crop testing, unless explicitly stated. With regard to the comparison of method complexity, we consider to use total FLOPs (FLOPs per video) as the main metric.

%% file: tex/5_2_main_results.tex
\subsection{Main Results}
\vspace{-4pt}
Here we present the main results of our \autotwostreamnetspace models in Table \ref{tab:main_res}, including S, M and L variants which stand for small, medium, and large model. We also compare them with other state-of-the-art video models whose architectures are searched. 
Auto-TSNet-S and Auto-TSNet-M shared the same architecture, and only differ in the input video spatial resolution ($182^2$ vs $256^2$). Auto-TSNet-L model is obtained by naively stretching the depth of Auto-TSNet-S by $2\times$, and increasing the input spatial resolution to $356^2$. We also report results of another three variants of our model, namely Auto-TSNet-S$^\dagger$, Auto-TSNet-M$^\dagger$ and Auto-TSNet-L$^\dagger$, where the attention blocks are removed. 

In the first section of Table \ref{tab:main_res} where we compare small models, Auto-TSNet-S$^\dagger$ outperforms X3D-S by a significant margin of \textbf{1.3\%}, while using similar FLOPs. When attention blocks are searched and added, Auto-TSNet-S further improves the accuracy by \textbf{0.8\%}. In the second section where we compare medium-size models, Auto-TSNet-M improves X3D-M by a large gap of \textbf{1.3\%} using similar FLOPs. The performance of Auto-TSNet-M is even on par with X3D-L ($77.3\%$ Vs $77.5\%$), while using \textbf{$66\%$} less FLOPs.
In the last section of Table \ref{tab:main_res} for comparing big models, Auto-TSNet-L significantly surpasses X3D-L by $1.4\%$ ($78.9\%$ Vs $77.5\%$), while using \textbf{20\%} less FLOPs. 

\begin{table}[t]
\vspace{-3pt}
\centering
\resizebox{0.5\textwidth}{!}{
  \begin{tabular}{c|c|c|c|c}
     \multirow{2}{*}{Model} & Params & GFLOPS & Total & Top-1  \\
      &(M) & $\times$views & GFLOPs & Acc ($\%$) \\ 
    \specialrule{.15em}{.1em}{.1em}
      X3D-S~\cite{x3d} & $3.8$ & $2.7$ $\times$ 30&81& $73.3^{1}$ \\
      Auto-TSNet-S$^\dagger$(ours) & $7.7$ & $2.8$ $\times$ $30$ &$84$& $74.6$ \\
      \textbf{Auto-TSNet-S (ours)} & $8.6$ & $3.4$ $\times$ $30$ &$102$& \textbf{75.4} \\ \hline
      
      EfficientNet3D-B3~\cite{2019efficientnet} & $8.2$ & $6.9$ $\times$ $10$ &$69$& $72.4$  \\
      X3D-M~\cite{x3d} & $3.8$ & $6.2$ $\times$ $30$ &$186$& $76.0$ \\
      Auto-TSNet-M$^\dagger$(ours) & $7.7$ & $5.2$ $\times$ $30$ &$156$& $76.6$\\
      \textbf{Auto-TSNet-M (ours)} & $8.6$ & $6.1$ $\times$ $30$ &$183$& \textbf{77.3}\\\hline
      
      EfficientNet3D-B4~\cite{2019efficientnet} & $12.2$ & $23.8$ $\times$ $10$  &$238$& $74.5$ \\
      X3D-L~\cite{x3d} & $6.1$ & $24.8$ $\times$ $30$ &$744$& $77.5$ \\
      Auto-TSNet-L$^\dagger$(ours) & $12.2$ & $18.1$ $\times$ $30$ &$543$& $78.3$\\
      \textbf{Auto-TSNet-L (ours)} & $13.2$ & $19.9$ $\times$ $30$ &$597$& \textbf{78.9}\\
     \bottomrule
  \end{tabular}
}
\vspace{-3pt}
  \caption{\textbf{Comparisons with other NAS models on Kinetics-400.} \atn and X3D models are evaluated using 10-LeftCenterRight testing. $\dagger$ denotes the model without attention blocks.}
\label{tab:main_res}
\vspace{-6pt}
\end{table}

% \begin{table}[t]
% \centering
% \resizebox{0.5\textwidth}{!}{
%   \begin{tabular}{c|c|c|c|c}
%      \multirow{2}{*}{Model} & Params & GFLOPS & Total & Top-1  \\
%       &(M) & $\times$views & GFLOPs & Acc ($\%$) \\ 
%     \specialrule{.15em}{.1em}{.1em}
%       X3D-S~\cite{x3d} & 3.8 & 2.7 $\times$ 30&81& $73.3^{1}$ \\
%       \textbf{Auto-TSNet-S$^\dagger$(ours)} & 7.7 & 2.8 $\times$ 30 &84& 74.7 \\
%       \textbf{Auto-TSNet-S (ours)} & 8.6 & 3.4 $\times$ 30 &102& \textbf{75.4} \\ \hline
%       EfficientNet3D-B3~\cite{2019efficientnet} & 8.2 & 6.9 $\times$ 10 &69& 72.4  \\
%       X3D-M~\cite{x3d} & 3.8 & 6.2 $\times$ 30 &186& 76.0 \\
%       \textbf{Auto-TSNet-M$^\dagger$(ours)} & 7.7 & 5.2 $\times$ 30 &156& \textbf{76.6}\\ \hline
%       EfficientNet3D-B4~\cite{2019efficientnet} & 12.2 & 23.8 $\times$ 10  &238& 74.5 \\
%       X3D-L~\cite{x3d} & 6.1 & 24.8 $\times$ 30 &744& \textbf{77.5} \\
%       \textbf{Auto-TSNet-M (ours)} & 8.6 & 6.1 $\times$ 30 &183& 77.3\\
%      \bottomrule
%   \end{tabular}
% }
% \vspace{-3pt}
%   \caption{\textbf{Comparisons with other NAS models on Kinetics-400.} \atn and X3D models are evaluated using 10-LeftCenterRight testing. $\dagger$ denotes the model without attention blocks.}
% \label{tab:main_res}
% \vspace{-6pt}
% \end{table}

\footnotetext[1]{Accuracy reported in the \href{https://github.com/facebookresearch/SlowFast}{official repo}. X3D paper doesn't report the result of X3D-S using 10-LeftCenterRight testing.} 

%% file: tex/5_3_twostream_baseline.tex
\subsection{From One-Stream to Two-Stream Model}
\label{sec:from_one_to_two_streams}

% \Heng{This section is not very relevant to the idea we want to sell. And it only talks about manually designed model, as listed in Table 4. A more interesting comparison could be "X3D" vs. "searched one stream model" vs. "searched two stream model". We can argue that the searched one stream model is better than X3D, since it explores more design choices. And search two stream model is better than search one stream, since two stream has much larger search space, and can leads to better models.} \Xinyu{Makes sense. Actually we've already done such single-stream backbone search, which is compared with X3D-s. However, the performance of the searched one is on par with X3D-s.}

\begin{table}[t]
% \vspace{-0.6cm}
\centering
% \small
% \tiny
% \scriptsize
\resizebox{0.33\textwidth}{!}{

  \begin{tabular}{c|c|c}
    \multirow{1}{*}{Model} & FLOPS Ratio $P$ & Top-1 Acc (\%) \\
    \specialrule{.15em}{.1em}{.1em}
    X3D-S & -  &  $72.9$  \\ \hline
    \multirow{3}{*}{Manual-TSNet}  & $85\%$   & $72.8$   \\
      & $70\%$   & $\textbf{73.2}$   \\
     & $55\%$   & $72.4$ \\ \bottomrule
  \end{tabular}
}
\vspace{-3pt}
\caption{\textbf{Comparing X3D-S model with our Manual-TSNet models on Kinetics-400.} All models use about 2.0G FLOPS. Ratio \textit{P} denotes the ratio of FLOPS used by the sparse stream. }
\label{tab:handcrafted_two_stream_model}
\vspace{-15pt}
\end{table}
X3D~\cite{x3d} is a family of single-stream models, and achieve good performance on standard benchmarks. However, we hypothesize two-stream models can achieve higher performance than single-stream models. Before automatically searching for two-stream models, we manually build two-stream baseline models and compare with X3D models.

We fix the number of input frames of \streamonelowercase and \streamtwolowercase stream to be 4 and 32. For each stream, we reuse the macro structure of the X3D-S model and only modify it by uniformly scaling down the feature channel at each block under the following constraints. \textit{(i)}: the total FLOPS of the two-stream model per crop is close to X3D-s's FLOPs. \textit{(ii)}: the FLOPS of the \streamonelowercase stream account for $P\%$ of the overall FLOPS, where $P$ is a hyper-parameter. Following the design in SlowFast~\cite{slowfast}, we use time-strided convolution as fusion block between 2 streams, which are placed uniformly along the network. We experiment with 3 choices of $P \in \{85, 70, 55\}$, and denote the resulting models as \textit{Manual-TSNet-$P\%$}. The results are shown in Table~\ref{tab:handcrafted_two_stream_model}. On Kinetics-400, the Manual-TSNet-$70\%$ model achieves the best performance, and improves the X3D-S model by $0.3\%$ top-1 accuracy using similar FLOPS. %However, we also find using an improper choice, such as $P=85\%$, will lead to a model with inferior performance. 

Note our hand-crafted Manual-TSNet models only represent a fairly sparse set of data points in our multi-variate search space, and are not expected to be optimal in that space. Nevertheless, Manual-TSNet-$70\%$ already outperforms the delicately designed X3D-S model, which encourages us to more extensively explore the space at finer granularity beyond simple uniform channel scaling.

%% file: tex/5_5_search_two_stream_model.tex
\subsection{Progressive Search of Two-Stream Models}
\label{sec:exp_search_two_stream_model}
\vspace{-5pt}
\subsubsection{Searching For \streamoneuppercase Stream}
\vspace{-5pt}
As described in Section~\ref{sec:progressive_search}, we adopt a progressive search process that starts with searching the architecture of the \streamonelowercase stream. 
%We adopt PARSEC search method, and sample $K$ architectures at each search step. 
We set the target FLOPS of the \streamonelowercase stream to $1.4G$ FLOPS based on the design of our manually explored two-stream models Manual-TSNet-70\% in Table~\ref{tab:handcrafted_two_stream_model}. The search of the \streamonelowercase stream takes 2.3 days.
The searched \streamonelowercase stream achieves $70.8\%$ Top-1 accuracy on Kinetics-400, with 1.39G FLOPs, as shown in Table \ref{tab:step_2_searched_two_stream_models}, which is a good starting point for our progressive search.

\vspace{-7pt}
\subsubsection{Searching For \streamtwouppercase Stream and Fusion Blocks}
\vspace{-5pt}
In the 2nd step of the progressive search, we fix the architecture of the \streamonelowercase stream from the previous step as well as the model weights of the supernet, and further search for the architecture of the \streamtwolowercase stream and TS-Fusion blocks. We set the target FLOPS of the overall two-stream model to 2.0G. The results are shown in Table~\ref{tab:step_2_searched_two_stream_models}. The discovered model includes \streamonelowercase stream, \streamtwolowercase stream, and fusion blocks. Compared with the searched \streamonelowercase stream network from the previous step, the searched two-stream model achieves a performance boost of $3.3\%$. The FLOPs of the searched model is 2.05 GFLOPs, which is close to the target 2 GFLOPs.
%, while only has a minor increase of $0.6G$ FLOPS in model computation.

\vspace{-7pt}
\subsubsection{Searching for Attention Blocks}
\vspace{-5pt}
In the final step of the progressive search, we search for the insertion location of the attention block, where we choose to use GloRe as an instance of attention block. We consider inserting GloRe blocks at stage 2, 3, and 4. We uniformly pick 6 locations at stage 2, 3, and 4 for each stream as the candidate attention locations, which leads to a space of $2^{6} \times 2^{6} = 4096$ choices in total. We set the target FLOPs to 2.5G. The search only takes 0.9 days and the results are shown in the last row of Table~\ref{tab:step_2_searched_two_stream_models}.

The final searched Auto-TSNet model chooses to insert two GloRe blocks to the sparse stream of stage 3 (see Figure \ref{fig:vis}), which improves the accuracy of the searched architecture of step 2 by $0.5\%$ ($74.6\%$ Vs $74.1\%$). The FLOPs of the searched architecture is also closed to our target FLOPs.
\begin{table}[t!]
% \vspace{-0.6cm}
\centering
% \small
% \tiny 
% \scriptsize
\resizebox{0.49\textwidth}{!}{
 
  \begin{tabular}{ccccc|c|ccc}
     \# & \streamoneuppercase & \streamtwouppercase & Fusion & Attention & Search & 1-View & Top-1  \\
    Streams & stream & stream & block & block & days& GFLOPs & Acc (\%)  \\ 
    \specialrule{.15em}{.1em}{.1em}
    1 & \checkmark &  &   &  & $2.3$ & $1.39$ & $70.8$ \\
    2 & \checkmark & \checkmark & \checkmark & & $1.6$ & $2.05$ & $74.1$ ($+3.3$)  \\
    2 & \checkmark & \checkmark & \checkmark & \checkmark& $0.9$ & $2.46$ & $74.6$ ($+0.5$)  \\ \bottomrule
% \specialrule{.1em}{.05em}{.05em}
  \end{tabular}
}
\vspace{-6pt}
\caption{\textbf{Results of progressive architecture search.} %Models are evaluated on Kinetics-400, and accuracy is obtained with 10-LeftCenterRight testing. %We use \textit{manual} and \textit{auto} to denote whether the architecture is manually designed or automatically searched.
}
\label{tab:step_2_searched_two_stream_models}
% \vspace{-1pt}
\end{table}

%% file: tex/5_6_component_ablation.tex
\subsection{Ablation of Searchable Components}
\begin{table}[t]
\vspace{-0.2cm}
\centering
% \small
% \tiny
% \scriptsize
\resizebox{0.49\textwidth}{!}{ 

  \begin{tabular}{c|c|c|c}
    Design Method & Search Days & 1-View GFLOPs & Top-1 Acc (\%) \\
    \specialrule{.15em}{.1em}{.1em}
    Manual & - &1.40 &$70.2$   \\
    Searched& 2.3 & 1.39 & $\textbf{70.8}$ \\ \bottomrule
  \end{tabular}
}
\vspace{-6pt}
\caption{\textbf{Ablation on backbone search on \kdata.}}
\label{tab:backbone_search}
\vspace{-6pt}
\end{table}
In this section, we consider isolating each searchable component (backbone, fusion, and attention), to verify the necessity of searching for their architectures. %Since the effectiveness of backbone search has already been verified, We focus on the remaining two searchable components: fusion and attention.
% \vspace{-3pt}

\noindent \textbf{Searching for Backbone Only.} In the first step of progressive search of Auto-TSNet, we searched for a \streamonelowercase stream, whose target FLOPs is set to $1.4$ GFLOPS. We directly compare the performance of the searched \streamonelowercase stream with the \streamonelowercase stream of Manual-TSNet-70\% in Table~\ref{tab:backbone_search}, which shows that the searched \streamonelowercase stream outperforms the hand-crafted one by $0.6\%$ top-1 accuracy ($70.2\%$ vs $70.8\%$).

\noindent \textbf{Searching for Fusion Only.} We conduct a toy study where we only search for the fusion location between two streams with pre-defined architectures, identical to those in Manual-TSNet-70\%. The candidate fusion operations include time-strided-conv and no-connection. We create 3 variant models based on the backbone architecture of Manual-TSNet-70\% with different number of fusion blocks, which are 0, 5 and 10 respectively. The candidate searchable fusion locations fully cover the fusion location of Manual-TSNet-70\% and its two variants. The results are in Table \ref{tab:connection_search}. We observe the searched architecture has surpassed other manually designed baselines with a considerable accuracy boost 0.5\%, demonstrating fusion search is non-trivial.

\begin{table}[t]
% \vspace{-0.6cm}
\centering
% \small
% \tiny
% \scriptsize
\resizebox{0.46\textwidth}{!}{ 

  \begin{tabular}{c|c|c|cccc|c}
    Design & Search  & Total Fusion &\multicolumn{4}{c|}{Fusion Blocks per Stage} & Top-1  \\
    Method & Days & Blocks &S1 & S2 & S3 & S4 & Acc (\%) \\
    \specialrule{.15em}{.1em}{.1em}
    \multirow{3}{*}{Manual}& \multirow{3}{*}{-} &0&  - & - & - & - &$70.7$   \\
     && 5&$+1$ & $+1$ & $+2$ & $+1$ & $73.2$  \\
     && 10&$+2$ & $+2$ & $+4$ & $+2$ & $71.2$  \\ \midrule
     Searched& 0.7 &4& $+1$ & $+1$ & $+2$ & - & $\textbf{73.7}$ \\\bottomrule
  \end{tabular}
}
\vspace{-6pt}
\caption{\textbf{Evaluating the choices of fusion locations on Manual-TSNet on \kdata.} }
\label{tab:connection_search}
\vspace{-10pt}
\end{table}

\begin{table}[t!]
\centering
\resizebox{0.49\textwidth}{!}{

  \begin{tabular}{c|c|c|ccc|ccc}
    Design &Search& Use & \multicolumn{3}{c|}{GloRe per Stage} & Params & 1-View & Top-1  \\
     Method &Days& GloRe & S2 & S3 & S4 & (M) & GFLOPs & Acc (\%)  \\ 
    \specialrule{.15em}{.1em}{.1em}
     Manual-TSNet-70\% & - & $\times$ & - & - & - & $7.71$ & $2.03$ & $73.2$ \\ \midrule
     \multirow{3}{*}{Manual} &\multirow{3}{*}{-}& \multirow{3}{*}{\checkmark} & $+2$ & - & - & $8.03$ & $2.19$ & $72.7$  \\
     &&  & $+2$ & $+2$ & - & $8.24$ & $2.41$ & $73.2$  \\
     &&  & $+2$ & $+2$ & $+2$ & $8.37$ & $2.51$ & $73.3$  \\ \midrule
    Searched & $0.9$ & \checkmark & +1 & +1 & - &  $8.15$ & $2.40$ & $\textbf{73.6}$ \\ \bottomrule
  \end{tabular}
}
\vspace{-6pt}
\caption{\textbf{Comparing manually inserted and searched attention blocks in Manual-TSNet-70$\%$ model on \kdata.}}
\label{tab:glore_search}
\vspace{-6pt}
\end{table}

\noindent \textbf{Searching for Attention Only.} To demonstrate the need of searching attention blocks, we also conduct an experiment where we only search for the location of attention block, while keeping other searchable components fixed (identical to Manual-TSNet-70\%). We adopt Manual-TSNet-70\% as the baseline, along with its variants with different insertions of attention blocks. Results in Table~\ref{tab:glore_search} confirm  the network with searched attention location outperforms other manual designed variations. Compared to Manual-TSNet-70\%, the searched network obtains an accuracy boost of $0.4\%$. 
% This demonstrates the location of attention blocks is worth being searched.

%% file: tex/5_6_progressive_ablation.tex
\subsection{Ablation of Progressive Search}
\vspace{-5pt}
In our paper, we adopted a progressive search algorithm to decompose the tremendous search space ($2\times10^{53}$) into several small sub search space ($[8\times10^{24}, 6\times10^{24}, 4.096\times10^{3}]$), for more efficient search. We compare the searching time of progressive search and one-step search, as well as their search performances. Here one-step search denotes searching all variates simultaneously, instead of searching part by part in a progressive way. The only difference between them is that the one-step search algorithm search for all variables at once, and we keep other settings to be identical for fairness. We also introduce the entropy of architecture parameters $\pmb{\alpha}$ as a convergence indicator for the search process. A lower entropy value indicates the searching is closer to convergence. The results are shown in Table \ref{tab:progressive}.
\begin{table}[t!]
\centering
\resizebox{\linewidth}{!}{
  \begin{tabular}{c|cc|c|c|c}
     Progressive & Search & End & Params& GFLOPs & Top-1  \\
      search & days & entropy &(M)& $\times$views & Acc ($\%$) \\ 
    \specialrule{.15em}{.1em}{.1em}
    $\times$ & 6.5 & 18.9 & 8.9 & $ 3.41\times30$ & 72.8  \\
      \checkmark & 4.8 & $9.7$ & 8.6 & $3.25\times30$& $75.4$ \\
     \bottomrule
  \end{tabular}
}\vspace{-6pt}
  \caption{\textbf{Comparisons of one-step search and progressive search on Kinetics-400.} Models are evaluated using 10-LeftCenterRight testing. The entropy of progressive search is averaged across the search for each part.}
\label{tab:progressive}
\vspace{-10pt}
\end{table}
We confirm the progressive search mechanism reduces the search time by $26\%$, and also converges better (18.9 vs 9.7 in entropy). The searched architecture of progressive search also achieves better performance, outperforming the non-progressive one by $2.6\%$.

%% file: tex/5_7_sota_models_kinetics.tex
\subsection{Comparisons with SOTA on \kdata}
\label{sec:compare_sota_kinetics}

\begin{table}[t!]
\vspace{-0.6cm}
\centering
% \small
% \tiny
% \scriptsize
\resizebox{0.51\textwidth}{!}{

  \begin{tabular}{c|l|cccc}
    \# & \multirow{2}{*}{Model} & Pre- & Params & Total & \multicolumn{1}{c}{Top-1}  \\
    streams & & training &  (M) & GFLOPs & Acc ($\%$) \\ 
    \specialrule{.15em}{.1em}{.1em}
    \multirow{7}{*}{1}& I3D~\cite{i3d} & \multirow{7}{*}{ \rotatebox[origin=c]{90}{ImageNet}  } & $12$  & - & $71.1$ \\
     & MF-Net~\cite{chen2018multi} &  & $8.0$&$555$& $72.8$ \\ 
     & TSM~\cite{tsm} &  & $24.3$ &$650$& $74.7$  \\
     & Nonlocal, R50~\cite{nonlocal_block} &  & $35.3$ &$8,460$& $76.5$ \\
     & Nonlocal, R101 &  & $54.3$ & $10,770$ & $77.7$ \\
    & Oct-I3D+NL &  & $33.6$ &$867$& $75.7$ \\
    & SmallBigNet~\cite{SmallBigNet} & & - &$5,016$& $77.4$ \\
    
     \hline
     
    \multirow{7}{*}{1}& R(2+1)D~\cite{r2plus1} & \multirow{7}{*}{-} & $63.6$ &$17,480$& $72.0$ \\
    & ip-CSN-152~\cite{csn} & & $32.8$ &$30,270$& $77.8$ \\
    & VoV3D-M~\cite{vov3d} & & $3.8$ &$132$& $73.9$ \\
    & VoV3D-L & & $6.2$ &$27$9& $76.3$ \\
    & X3D-S~\cite{x3d} & & $3.8$  &$75$& $73.3^1$ \\
    & X3D-M & & $3.8$ &$186$& $76.0$ \\
    & X3D-L & & $6.1$ &$744$& $77.5$ \\
    & X3D-XL & & $11.0$ &$1,452$& $79.1$ \\
     
     \hline
      
    \multirow{4}{*}{2} & 2-Stream I3D~\cite{i3d} & \multirow{4}{*}{ \rotatebox[origin=c]{90}{ImageNet} } & 25 & - & 75.7 \\
    & 2-Stream S3D-G\cite{s3d} &  & $23.1$ & - & $77.2$  \\
    & 2-Stream TSN~\cite{temporalsegmentnetwork} &  &  - &-& $73.9$  \\
    & 2-Stream ARTNet, R18~\cite{ARTNet} & & - & - & $72.4$ \\
    
    \hline
    
    \multirow{6}{*}{2} & 2-Stream R(2+1)D\cite{r2plus1} & \multirow{6}{*}{-} & $127.2$ &$34,960$& $73.9$ \\
    & A3D-SF 4$\times$16, R50~\cite{a3d} & & $34.4$ & $1,083$ & $75.7$  \\
    & SF 4$\times$16, R50~\cite{slowfast} & & $34.4$  &$1,083$& $75.6$  \\
    & SF 8$\times$8, R101 & & $53.7$  &$3,180$& $77.9$  \\
    & SF 16$\times$8, R101 & & $53.7$  &$6,390$& $78.9$ \\ 
    % & SF 8$\times$8, R101+NL & & 59.9 &3,480& 78.7 \\
    % & SF 16$\times$8, R101+NL & & 59.9  &7,020& 79.8 \\ 
    \hline
%     & \textbf{Manual-TSNet-S (ours)} & & - &  2.1$\times$30 & 74.02 & \textbf{-} \\
%     & \textbf{Manual-TSNet-M (ours)} & & - &  3.98$\times$30 & 76.24 & \textbf{-} \\
    %  \multirow{6}{*}{2} & Auto-TSNet-S$^\dagger$(ours) & \multirow{6}{*}{-} & 7.7 &84.9& 74.1 \\
    \multirow{3}{*}{2}  & \textbf{Auto-TSNet-S (ours)} & \multirow{3}{*}{-}  & $8.6$ &$102$& $\textbf{75.4}$ \\
    %  & Auto-TSNet-M$^\dagger$(ours) & & 7.7 &156& 76.1 \\
    & \textbf{Auto-TSNet-M (ours)} & & $8.6$ & $183$ & $\textbf{77.3}$\\
    % & Auto-TSNet-L$^\dagger$(ours) & & 12.2 &543& - \\
    & \textbf{Auto-TSNet-L (ours)} & & $13.2$ & $597$ & $\textbf{78.9}$ \\
\bottomrule
  \end{tabular}
}
\vspace{-6pt}
\caption{\textbf{Comparing \autotwostreamnetspace models with others on \kdata.} }
\label{tab:sota_comparison_kinetics}
\vspace{-0.4cm}
\end{table}

In Table~\ref{tab:sota_comparison_kinetics}, we compare Auto-TSNet models with state-of-the-art results using \textit{10-LeftCenterRight} testing setting. 
%Compared to big models like SF $4\times16$ R50, Auto-TSNe-M outperforms it with a significant gain of \textbf{1.6\%}, while the FLOPs cost is \textbf{6.2}$\times$ smaller.
Compared with X3D-S, Auto-TSNet-S outperforms it with a significant margin of \textbf{2.4\%}. Using similar FLOPs, Auto-TSNet-M surpasses X3D-M by \textbf{1.3\%}. The performance of Auto-TSNet-M is even on par with X3D-L, while using \textbf{$66\%$} less FLOPs. Auto-TSNet-L surpasses X3D-L by a considerable gap of \textbf{1.4\%}, while using \textbf{20\%} fewer FLOPs. The performance of Auto-TSNet-L is even close to that of X3D-XL (78.9\% vs 79.1\%), but saves FLOPS by $60\%$. Comparing with SF $16\times8$ R101, Auto-TSNet-L achieves the same performance with it, but it is incredibly \textbf{11}$\times$ smaller than SF $16\times8$ R101 w.r.t. FLOPs cost.

%Auto-TSNet-L surpasses X3D-L by $0.5\%$, while has $20\%$ fewer FLOPs.

%% file: tex/5_8_sota_models_sth_sth.tex
\subsection{Transferability on Something-Something-V2} 
% \Heng{Maybe a better title: Generalize Auto-TSNet to different datasets. We should mention that our method is a general framework, and can adapt to different datasets, and discover new architectures automatically. Mention the changes we made to make it work on Something-Something-V2 if any. Also it may worth to compare the difference of the architecture learned from Kinetics and Something-something. Highlight those difference to backup the argument that NAS is necessary as different datasets has different characteristics, and our method can learn new architectures that are better fit to each dataset.}
\label{sec:compare_sota_sth_sth}

The next question we try to answer is: \textit{Does the searched \autotwostreamnetspace models overfit the dataset?} In other words, would the same model attain high performance, when we train it from scratch on a different dataset?  To address this question, We further evaluate the searched \autotwostreamnetspace models on Something-Something-V2 dataset. As mentioned before, the characteristic of  Something-Something-V2 dataset is quite distinct from \kdata, which focuses more on temporal modeling. The results are shown in Table~\ref{tab:sota_comparison_sth_sth_v2}, where we choose state-of-the-art efficient models under 50 GFLOPs to compare. Auto-TSNet models achieve highly competitive performance \textbf{without any pre-training}.
Comparing with GST$_{8F}$, Auto-TSNet-S shows a significant performance gain of $0.7\%$, while using 20\% fewer FLOPs. Auto-TSNet-M further boosts the accuracy to $63.6\%$,  and outperform all other methods.

%Comparing with MSTNet, the proposed Auto-TSNet-S provides better performance (62.3\% vs 59.5\%), while only has 67\% of MSTNet's FLOPs cost. When compared with TANet and bLVNet-TAM, the Auto-TSNet-M surpasses them with a considerable margin of $0.8\%$ and $1.1\%$, while only taks $1.4\times$ more FLOPs cost.

%We still find the discovered \autotwostreamnetspace maintains strong ATC trade-off. For example, compared to recent models such as $\text{SmallBigNet}_{8F}$, Auto-TSNet-M achieves $0.5\%$ better accuracy, while requiring $12\times$ fewer FLOPs.

% We also evaluate our searched \autotwostreamnet models on Something-Something-V2 dataset, and compare the results with SOTA methods in Table~\ref{tab:sota_comparison_sth_sth_v2}. As mentioned before, the characteristic of  Something-Something-V2 dataset is quite distinct from Kinetics-400, which  focuses more on temporal modeling. 

% \zhicheng{Based on the final results, compare Auto-TSNet results with SOTA results. BALABALA.}

% zyan3: add a dummy box to take the space which is expected to discuss results in tab:sota_comparison_kinetics
% \begin{solutionorbox}
% \vspace{8cm}
% \end{solutionorbox}

%% file: tex/5_9_analysis_of_model.tex
\subsection{Visualization of \autotwostreamnetspace Model}
\begin{figure}[t!]
\begin{center}
 \includegraphics[width=1.0\linewidth]{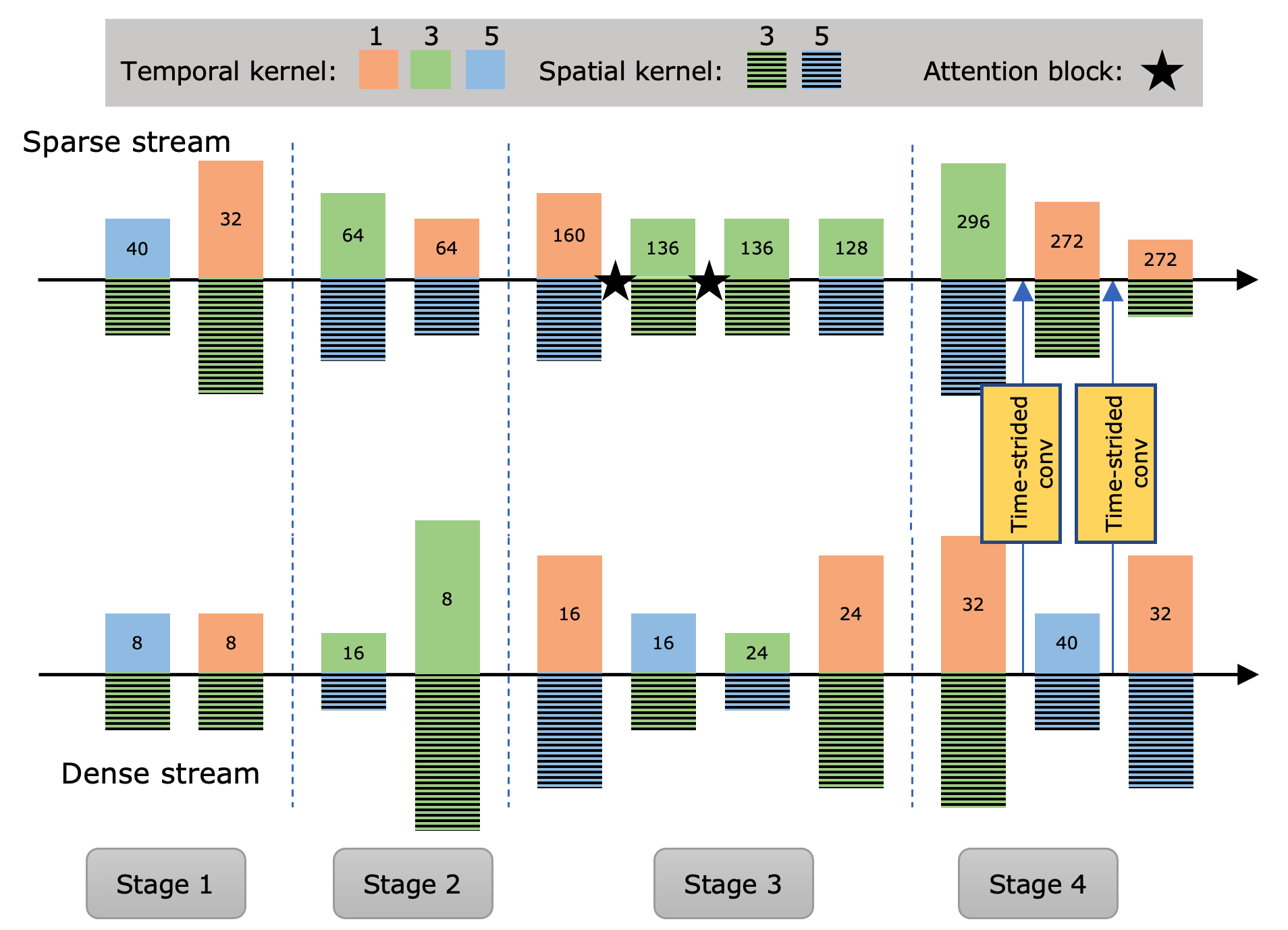}
\end{center}
\vspace{-1.5em}
  \caption{ \textbf{The visualization of our searched Auto-TSNet-S}. Each rectangle box (except fusion blocks between the two streams) denotes a block group. We use height to denote the expansion rate, and the block output channel of the group is marked in the box. Different colors are used to denote kernel size, as shown in the legend. We use the shadow texture to distinguish spatial kernel size from temporal one. The star symbol is used to represent the attention block.
  %The architecture visualization of our searched Auto-TSNet. Each box in the figure denotes a group of inverted residual blocks. Different color denotes the different kernel size of the depth-wise convolution in the inverted residual block, as depicted in the bottom of the figure. The number in the box means the output channel number, and the width of a box represents the expansion ratio of the inverted residual block. The lateral arrow represents the connection type, and the number in the arrow means the channel expansion ratio in the connection operation.
  }
\label{fig:vis}
\vspace{-10pt}
\end{figure}

\begin{table}[t!]
\vspace{-0.3cm}
\centering
% \small
% \tiny
% \scriptsize
\resizebox{0.5\textwidth}{!}{

  \begin{tabular}{c|l|cccc}
    \# & \multirow{2}{*}{Model} & Pre- & Params  &Total& \multicolumn{1}{c}{Top-1}  \\
    streams & & training &  (M) & GFLOPs & Acc ($\%$)  \\ 
    \specialrule{.15em}{.1em}{.1em}
    \multirow{12}{*}{1}& TSN~\cite{temporalsegmentnetwork} & \multirow{12}{*}{ \rotatebox[origin=c]{90}{ImageNet}   } & - &-& 41.1 \\
    & TSN, Dual attention~\cite{dual_attention} & & -  &-& $42.1$ \\
      & TRN, Dual attention & & &-& $51.6$  \\
    & TRN~\cite{zhou2018temporal} & & - & - & $48.8$ \\
    %  & TSM_{8F}~\cite{tsm} &  & 23.9 &197& 59.1  \\
    %  & TSM_{16F} &  & 23.9 &394& 63.4 \\
     & TSM, Dual attention&  & - &-& $55.0$  \\
    %  & TSM + TPN~\cite{TPN} & & - & - $\times$1 & 62.0 & - \\
    %  & STM$_{8F}$~\cite{stm} & & 24.0  &66.6& 62.3  \\
    %  & STM$_{16F}$ & & 24.0 &133& 64.2  \\
    %  & GST$_{16F}$ & & -  &59& 62.6\\
     & I3D + STIN + OIE~\cite{STIN} & & -  &-& $60.2$  \\
     & Dynamic Inference~\cite{wu2020dynamic} & & -&$35.4$& $58.2$  \\
    %  & TANet~\cite{TANet} & & 25.6  &516& 64.6  \\
    %  & SmallBigNet_{8F}~\cite{SmallBigNet} & &  &312& 61.6  \\
    %  & SmallBigNet_{16F} & &  &630& 63.8 \\
     & bLVNet-TAM~\cite{fan2019more} &  & $40.2$ & $32.1$ & $60.2$ \\
     & MSTNet~\cite{shan2020mixtconv} & & $24.3$ & $33.2$ & $59.5$ \\
     & TANet~\cite{liu2020tam} & & $30.4$ &$33.0$ & $60.5$ \\
          & GST$_{8F}$~\cite{GST} & & - &$29.5$& $61.6$  \\
    %  & MSNet~\cite{MSNet} & & 49.2  &1,010& 67.1 \\
    %  & TDRL~\cite{TDRL} & & -  &99& 65.1 \\
    %  & STF~\cite{STF} & & & & & \\
    %  & GSM~\cite{GSM} & & & & & \\
    %  & TEA~\cite{TEA} & & - & -$\times$3 & 63.2 & 89.7 \\
     & TEINet~\cite{TEINet} & & - & $33.0$ & $61.3$ \\
    %  & VSN~\cite{VSN} & & 85.8 & - & 66.1 & - \\
     \hline
      
    \multirow{2}{*}{2} & 2-Stream TRN~\cite{zhou2018temporal} & \multirow{2}{*}{ ImageNet } & - & - & $55.5$ \\
    & 2-stream TRN, Dual attention~\cite{dual_attention} & &- &- & $58.4$ \\
    % & ABM-C-in, R34~\cite{ABM} & & & - & 63.9  \\
    % & 2-Stream TSM~\cite{tsm} &  & 47.8 & 286 &66.0 \\

    \hline
    
    \multirow{2}{*}{2} & \textbf{Auto-TSNet-S (ours)} & \multirow{2}{*}{-} & 8.6 & 23.7& \textbf{62.3} \\
    & \textbf{Auto-TSNet-M (ours)} & & 8.6 &46.5& \textbf{63.6}  \\ \bottomrule
    % & \textbf{Auto-TSNet-L (ours)} & & 6.8 &  12.4$\times$10 & \textbf{TBD} & \textbf{TBD} \\
  \end{tabular}
}
\vspace{-2pt}
\caption{\textbf{Comparing \autotwostreamnetspace models with other efficient SOTA models (under 50 GFLOPs) on Something-Something-V2 dataset.} }
\label{tab:sota_comparison_sth_sth_v2}
\vspace{-6pt}
\end{table}
\vspace{-1pt}
We visualize the discovered Auto-TSNet architecture in Figure~\ref{fig:vis}. For backbone, we can observe that the spatial kernel size of 5 and temporal kernel size of 1 are widely used in both streams, which is quite different from the design in current hand-crafted models. Surprisingly, we notice that the attention blocks only appear on the sparse stream, indicating they are less effective on the dense stream. Unlike manual designed architectures which often favor the uniform choices of search variables, the searched model uses nonuniform architecture choices.

% \vfill

% zyan3: add a dummy box to take the space which is expected to visualize model architectures.
% \begin{solutionorbox}
% \vspace{6cm}
% \end{solutionorbox}

%% file: tex/6_conclusion.tex
\section{Conclusions}
\vspace{-3pt}
We present an approach to searching for high-performance \twostream models for video recognition. We meticulously prescribe a multivariate space with 6 search variables, which have a substantial impact on the model performance and complexity, and reflect the large variations in designing \twostream models. A progressive search process is proposed to efficiently search in the proposed large design space, and the discovered \twostream models outperform other models on two large-scale action recognition benchmarks.